# Neural Computing

Ayushe Gangal[1], Peeyush Kumar[1], Sunita Kumari[1], Aditya Kumar[2]
G.B. Pant Government Engineering College[1], Deenbandhu Chhotu Ram University Of Science And Technology[2]
ayushe17@gmail.com, peeyushstark@gmail.com, sunita2009@gmail.com, adityakumar333stark@gmail.com

## ABSTRACT

This chapter aims to provide next level understanding of the problems of the world and the solutions available to those problems, which lie very well within the domain of neural computing, and at the same time are intelligent in their approach, to invoke a sense of innovation among the educationalists, researchers, academic professionals, students and people concerned, by highlighting the work done by major researchers and innovators in this field and thus, encouraging the readers to develop newer and more advanced techniques for the same. By means of this chapter, the societal problems are discussed and various solutions are also given by means of the theories presented and researches done so far. Different types of neural networks discovered so far and applications of some of those neural networks are focused on, apart from their theoretical understanding, the working and core concepts involved in the applications.

## INTRODUCTION

The brain is the most intricate organ in the entire human body. It is responsible for our every thought, action, feeling and memory. The average human brain consists of 100 billion neurons and each neuron is connected to 10,000 other neurons (Richard, 2018). The initial inspiration of the idea behind neural networks loosely mimics the biological neurons present in the human brain to varying extents. The glorious ambition of neural networks is to build machines with computational power comparable to the computational power of the human brain (Kurenkov, 2015). But this is an extremely arduous and challenging task, provided that the computational power of the fastest computer present today is a minute fragment of the computational power of the human brain. The research on neural networks comprises of two major known goals, namely, understanding the human brain, and emulating the brain (Calimera, Macii & Poncino, 2013). The former goal is related to neuroscience, a branch of science concerned with the structure and function of the nervous system and the brain, and the latter is chiefly applied in nature and is associated with computer science and artificial intelligence, a sub-field of computer science. Neural computing ("Neural computation", 2019) is the processing of hypothetical information by neurons, forming a network.

But why were neural networks developed in the first place when digital computers were already present? Though the traditional digital computer was able to perform the desired computations well and fast, it could do what it was told to do and how it was told to do the task (Copeland, 2019), but the traditional digital computers could not help us or themselves, for that matter, if the problem to be solved was not fully understood by us. Furthermore, as artificial neural networks could learn on their own, they are better equipped than any standard computational algorithm, to make use of the noisy and incomplete data that is usually available to solve real world problems. Figure 1 depicts various applications of neural computing in today's world.

By the means of this chapter, the authors aim to explore the far-reaching field of neural networks, in a complete, concise and cogent manner, and to discuss the theory supporting the concept of neural networks, their origin, the idea behind the discovery made and the different types of major neural networks discovered so far, to elucidate the readers on the latest and groundbreaking research in the diverse and arcane field of artificial intelligence, that is, neural networks. Numerous important applications of neural networks from day-to-day life and lifestyle enhancing technological inventions are also highlighted in a well-grounded and lucid manner.

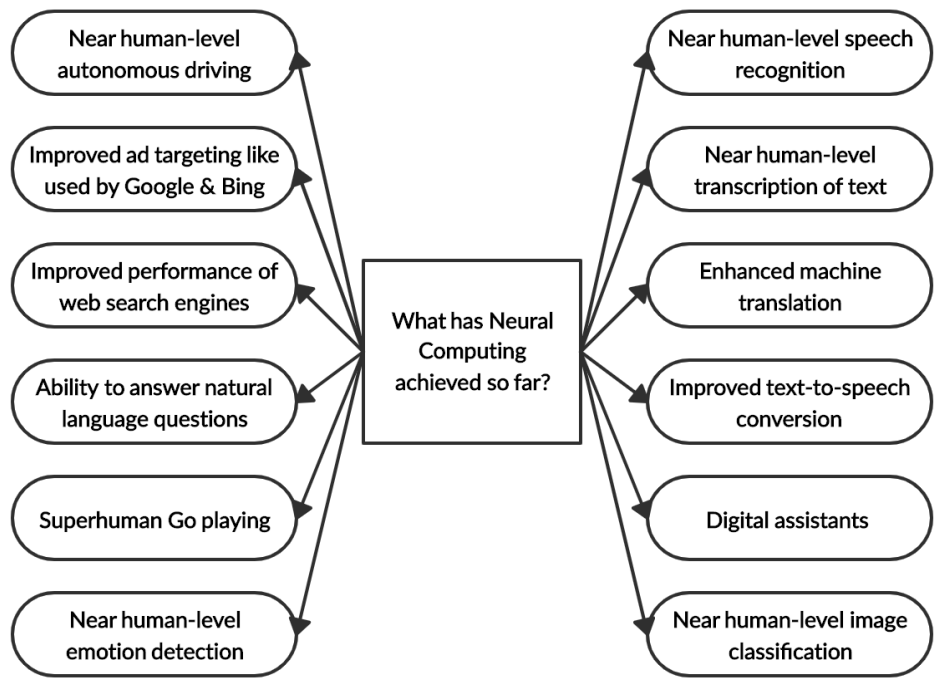

*Figure 1. Various applications of neural computing*

**Roadmap**

The structure of this chapter is fairly simple. Starting from the earliest research done, the successes, failures and impacts of these findings on the world as a whole in section 1, and the problems conquered and the solutions offered by the neural networks in section 2. Moving on to elucidating the core concept and requirements that evoked the need for the creation of neural networks in section 3. Section 4 is dedicated to discuss the various types of neural networks that have been discovered and are majorly used, or have potential to be majorly used for a wide range of world/societal problems in a brief and informative manner. Major day-to-day applications regarding the types of neural networks are briefly highlighted in section 4. The various types of neural networks which focussed on are, Convolutional Neural Networks, Deep Residual Network, Echo State Networks, Extreme Learning Machine, Liquid State Machine, Deep Convolutional Network, Deconvolutional Network, Radial Basis Function, Recurrent Neural Network, Long/Short term Memory, Auto Encoders and Neural Turing machines. The supporting labelled diagrams along with appropriate research are provided to enhance the reader's understanding of the applications provided. Section 5 offers a clear comparative analysis in the form of a tabular summarization of the neural networks discussed at length in this chapter. Section 6 talks about the conclusion and summarizes the content of the chapter for the readers' convenience.

**EVOLUTION OF NEURAL NETWORKS**

The tussle to obtain greater understanding of the human brain and its components, that is , the neurons has been on for a long time. (McCulloch & Pitts, 1943), a neuroscientist and a logician respectively, published a research paper called "A Logical Calculus of the Ideas Immanent in Nervous Activity". Their research dealt with the deeper understanding of the collection of cells, called neurons, working collectively for the brain to function and perform convoluted tasks. This acted as a major breakthrough and the commencement of research in this field. The MCP neuron, short for McCulloch and Pitts neuron, significantly assisted in the development of artificial neural networks, as it is a neural network which functions with biological neurons as its base.

Alan Turing wrote a report in 1948 stating (Norman, n.d.) that an intelligent machine has a mind of a child, which can be used to learn things on its own and thus made intelligent on its own, rather than an adult mind filled with pre existing ideas and notions. He also proposed a test called the Turing test, which could be used to classify a machine as intelligent or not.

Though the MCP model was a breakthrough, it had some limitations, which were conquered by the single layer perceptron, which can learn from data, by (Rosenblatt, 1958). The Hebbian Theory of Synaptic Plasticity, which deals with the learning or adaptation of brain cells during the process of learning, formed the basis of this discovery. A supervised learning algorithm was proposed for the modified version of MCP neuron. It facilitated the artificial neural network to learn and train itself using the data by figuring out the correct weights during the learning process.

The usage of back propagation for the training of artificial neural networks was given by (Werbos, 1974). He was also the avant-garde of recurrent neural networks, a class of artificial neural networks which learn from training like the usual artificial neural network and remember data learnt from previous inputs while producing outputs.

The development and innovation has picked up pace since then. With the innovation of Boltzmann Machine by (Hinton & Sejnowski, 1985) and Max-pooling ("Max-pooling / Pooling"), which was introduced in 1992. Max pooling also aided in the recognition of 3D objects. Bidirectional Recurrent Neural Networks were introduced by (Schuster & Paliwal, 1997). Backpropagation along with max-pooling was able to significantly improve the performance of all previously introduced artificial neural networks. (Goodfellow, Pouget-Abadie, Mirza, Xu, Warde-Farley, Ozair, Courville & Bengio, 2014) introduced the Generative Adversarial Network (GAN) dominantly used for unsupervised learning. GAN works by employing two sets of neural networks competing against each other to analyze and capture every possible variation and interpretation of the dataset. Capsule Neural Networks were introduced by (Hinton & Sabour, 2017), and it is an approach to mimic the human biological neurons with much greater intensity. It is based on the addition of structures called capsules to the conventional Convoluted Neural Network (CNN). Though, there are variations among the different neural networks discovered, there is uniformity among the basic structure and function of all neural networks.

## ARE NEURAL NETWORKS THE NEW PANACEA?

Neural networks are looked upon to solve the complex real-life problems. They have the ability to learn from complex non-linear data by using input-output relationships, recognizing patterns, generalizing and inferencing, and making predictions. Neural networks are used in diversified fields like finance, transportation, natural language processing, medical and diagnostics, marketing, automation of vehicles, etc. These have proved to be strikingly effective where conventional formal analysis is difficult or has failed completely. Their strength lies in their expertise to process and make sense out of noisy and complex data.

Neural networks are being used to disentangle many real-world situations in the field of engineering, industries and business applications as well. In today's world, the companies and agencies which use or provide automation are running the gamut. Flight system control, power plants control, medical predictive systems, chemical plants, and other systems which can be automated, are being automated using neural networks. Quick manufacturing and quick retailing simultaneously have changed how the world works. Artificial intelligence has augmented sectors like banking, communications and media, consumer goods, defense and security, healthcare, manufacturing, public dealing, travel and hotels, and mining using neural networks as its base technology. Making the decisive power of machines more reliable is a continuous undertaking, and will likely remain one in the future as well. But neural networks induced automation has become a 'need' of the people. Major real-life applications aspire to enhance the quality of life of the people and therefore, minimize any type of loss.

In the colossal field of engineering, neural networks are used in aerospace for aircraft component fault detection, path simulations, auto-piloting and simulations, product design analysis, dynamic modeling, process and product diagnosis, inspection systems and  planning systems in manufacturing, virtual sensors, enhanced navigation systems, self driving cars, buses, trains, boats and ships in automotives. Credit and loan evaluation, currency and fraud risk prediction, loan and mortgage advising, customer mood analysis, market risk and research, demand-supply analysis, facial recognition, targeted detection and weapon simulation, signal processing and image denoising, disease prognosis using numerical, audio and pictorial data, hospital system's quality improvement and optimization, personalized prosthesis designing and optimized surgery and treatment plans devising, are some of the current applications of neural networks in the fields of banking, business analytics, defense, medicine and transportations.

Before the introduction of neural networks in these fields, all these systems were managed and operated by humans or conventional computers that needed orders for every task they performed. It was extremely time consuming and a trade off between the accuracy of the results and the time taken had to be maintained. With the introduction of neural networks, higher accuracies could be achieved, which lead to specifically defense and medical fields to flourish. One important thing to note is that, though neural networks proceed, conventional computers, neural networks and conventional algorithmic computers complement each other in reality. There are multiple tasks which may only require the computational power of the conventional computer. For most cases, both of these machines are used to provide higher efficiency and speed in the accomplishment of the task. Though right now, artificial intelligence and neural networks may seem like the 'conquer-all' tool. But as Richard Yonck, Founder and Lead Futurist of Intelligent Future Consulting said, that "everything looks like a nail, when your only tool is a hammer", but everything isn't, meaning not all problems can be conquered by deep learning. There are many problems which are yet to be taken down, like contextual language processing, engines for common sense and emotions and on-shot learning. These are some of the developments which might take a few decades to come to their full active form.

## BUILDING BLOCKS OF A NEURAL NETWORK

Machine learning is mainly about mapping of inputs to targets, and which is done via analyzing many training examples and their respective targets. In Neural computing, this mapping is done via deep or multiple sequences of data transformation and the degree to which this transformation should be done is Learned by exposure to the dataset. A fundamental way by which a neural network learns from a dataset, is given below. The whole  process is depicted in figure 2.

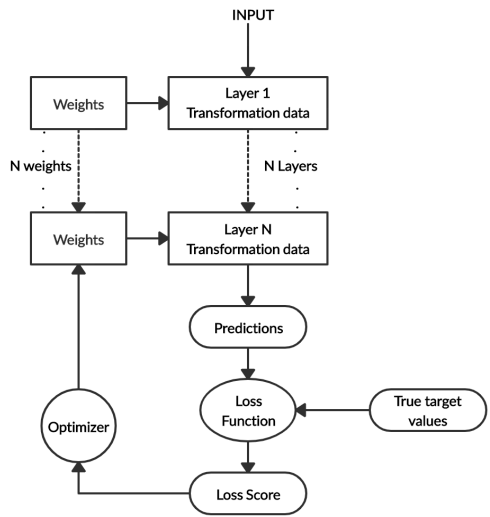

*Figure 2. Basic anatomy of a neural network*

1. A neural network contains many layers and the degree to which a layer transforms a given input is stored in the layer's weights, which is actually a set of numbers. So, an appropriate set of weights for all layers have to be learned in such a way that the neural network model maps the inputted examples correctly, to their associated targets. Initially, all the weights of the network are assigned a random value. In other words, a layer is a data processing module that takes an input and performs data transformation and provides a more meaningful representation of the inputted data as output. This data transformation is done in each layer with the help of an activation function. Activation functions are crucial for a neural net to learn and make sense of something convoluted and it also introduces non-linear properties to the neural net. The various types of activation functions are provided in table 1.

*Table 1. Table listing various activation functions*

| S.No. | Name of Activation Function | Description | Graph |
|---|---|---|---|
| 1 | Sigmoid Activation Function ("Sigmoid function") | The function f(x) is given by: $$f(x) = \frac{1}{1+e^{-x}}$$ It maps the input between the range 0 to 1. | *(sigmoid curve, range 0 to 1)* |
| 2 | Hyperbolic Tangent Function - Tanh ("Hyperbolic function") | The function f(x) is given by: $$f(x) = \frac{1-e^{-2x}}{1+e^{-2x}}$$ It maps the input between the range of -1 to 1. | *(tanh curve, range -1 to 1)* |
| 3 | ReLu - Rectified Linear Units Function (Nair & Hinton, 2010) | The function f(x) is given by: $f(x) = max(x,0)$ It gives either x as an output or 0. It has many advanced types like : 1. PReLu- Polynomial Relu 2. RReLu- Randomized Relu 3. LReLu- Leaky Relu 4. BReLu- Bipolar Relu 5. TReLu- Threshold Relu | *(ReLU curve)* |
| 4 | Step Function ("Step function") | The function f(x) is given by: $f(x) = \{1 \text{ if } x \geq 0 \text{ \& } 0 \text{ if } x < 0 \}$ It gives either 1 as an output or 0. | *(step function graph)* |

| 5 | Softmax Function ("Softmax function") | The function f(x) is given by: $$f(y=j|\theta^i) = \frac{e^{\theta^i}}{\sum_{j=0}^{k} e^{\theta^j}_k}$$ Where $\theta = W_0 X_0 + W_1 X_1 + .. + W_k X_k$ x = features and W = weights<br><br>It gives the output between 0 and 1, just like a sigmoid function. But it also divides each output such that the total sum of the outputs is equal to 1. | 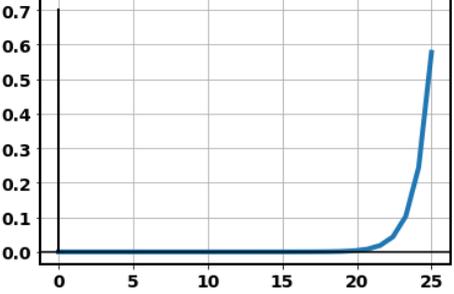 |
| --- | --- | --- | --- |
| 6 | GeLu - Gaussian Error Linear Unit (Hendrycks & Gimpel, 2016) | The function f(x) is given by: $$f(x) = 0.5x(1 + tanh(\sqrt{\tfrac{2}{\pi}}(x + 0.044715x^3)))$$ It is a state of the art activation function for Natural Language Processing. | 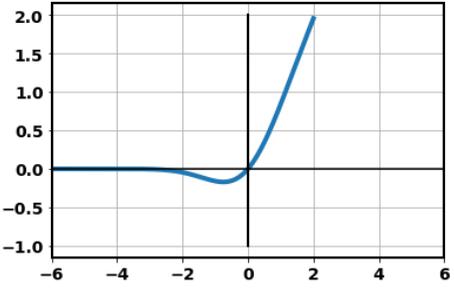 |

2. Learning a perfect combination of numbers for each layer (called weights) is an intimidating task, as there can be millions of weights and a slight change in each weight can affect the whole network. This daunting job of learning is done by a Loss function.

3. The loss function takes the predictions provided by the network and true target values, and by comparing them, computes a distance score (computes by how each prediction value differs from actual value). The few important types of loss functions are provided in Table 2. Where Y : Actual Value; y = Predicted Value; n = Number of Data points.

*Table 2. Table listing various loss functions*

| S.No. | Types of Loss functions | Description |
| --- | --- | --- |
| 1 | Mean Squared Error (MSE) ("Mean squared error") | MSE measures the average squared difference between the predicted value and the actual value. It is used for regression problems and is given by: $$MSE = \tfrac{1}{n}\sum_{i=0}^{n}(Y_i - y_i)^2$$ |
| 2 | Mean absolute Error (MAE) ("Mean absolute error") | MAE is a measure of difference between two continuous variables. It is given by: $MAE = \frac{\sum_{i=1}^{n}|Y_i - y_i|}{n}$ |
| 3 | Mean absolute Percentage Deviation (MAPD) ("Mean absolute percentage error", 2019) | MAPD is percentage measure of prediction accuracy of a forecasting method in statistics. It is given by: $$MAPD = \tfrac{100}{n}\sum_{i=1}^{n}\tfrac{|Y_i - y_i|}{|Y_i|}$$ |
| 4 | Mean Squared Logarithmic Error (MSLE) ("Mean squared logarithmic error loss function: Peltarion") | MSLE is given by: $$MSLE = \sqrt{\tfrac{1}{n}\sum_{i=1}^{n}(log(y_i + 1) - log(Y_i + 1))^2}$$ |

| 5 | Categorical Crossentropy (CC) ("Categorical crossentropy loss function: Peltarion Platform") | CC is used in neural computing for many class classification problems. It is given by: $$CC = \frac{-1}{n} \sum_{i=1}^{n} \sum_{c=1}^{C} 1_{y_i \varepsilon C_c} \log p_{model}[y_i \varepsilon C_c]$$ Where C : Total number of Classes ; pmodel : probability predicted by the model for the $i^{th}$ observation to belong to the $c^{th}$ category. |
|---|---|---|
| 6 | Binary Crossentropy (BC) ("Binary crossentropy loss function: Peltarion Platform") | BC is used in neural computing for two class classification problems. It is given by: $$BC = \frac{-1}{n} \sum_{i=1}^{n} [Y_i \log y_i + (1 - Y_i) \log(1 - y_i)]$$ |

4. The basic concept of neural computing is to use distance score in the form of a feedback signal to learn or adjust the weights for each layer, so that the score gets minimized for the training examples.

5. The adjustment of weights is done using an optimizer, which uses a central algorithm, used in neural computing called Backpropagation algorithm. Choosing the right optimizer for the right problem is crucial. The few important types of optimizers are provided in Table 3. Where j(w) : Loss function ; $\nabla j(w)$ : Gradient of loss function with respect to respective weights; w: weights; α = Learning rate.

*Table 3. Table listing various important optimizers*

| S.No. | Name of Optimizer | Description |
|---|---|---|
| 1 | Gradient Descent optimizer (GDO) (Cauchy, 1847) | In GDO each parameter w is changed using an algorithm given by: Algorithm : $w = w - \alpha \nabla j(w)$ It is a highly used and most basic algorithm for optimization of parameters in neural nets. The direction of increase is indicated by a gradient and the parameters are updated in the negative direction of gradient (or in the opposite direction of the gradient) to minimize the loss. |
| 2 | Stochastic Gradient Descent optimizer (SGDO) (Robbins & Monro, 1951) | In SGDO each parameter w is changed using an algorithm given by: Algorithm : $w = w - \alpha \nabla j(w; x(i); y(i))$ , where {x(i), y(i) are the training example} In SGDO model parameters are updated more frequently. The neural network model parameters are adjusted after loss computation on each training example unlike GDO, where weights are altered only one time. |
| 3 | Momentum Gradient Descent Optimizer (MGDO) (Qian, 1999) | In MGDO each parameter w is changed using an algorithm given by: Algorithm : $V(t) = \gamma V(t-1) + \alpha \nabla j(w)$ $w = w - V(t)$ Where γ : momentum MGDO was invented for softening the convergence and reducing the high variance in SGDO. It converges after SGDO. |
| 4 | Nesterov Accelerated Gradient Optimizer (NAGO) (Nesterov, 1983) | In NAGO each parameter w is changed using an algorithm given by: Algorithm : $V(t) = \gamma V(t-1) + \alpha \nabla j(w - \gamma V(t-1))$ $w = w - V(t)$ Where γ : momentum In MGDO, if the momentum is too high than the algorithm may miss the local minima, so to tackle this problem NAGO was developed. |

| 5 | Adagrad Optimiser (Duchi, Hazan & Singer, 2011) | In Adagrad optimizer each parameter w is changed using an algorithm given by:<br>Algorithm : $w_{t+1} = w_t - \dfrac{\alpha}{\sqrt{g_t + \varepsilon}}$<br>Where t : Time step ; $g_t$ : Sum of the squares of the past Gradients with respect to all weights; $\varepsilon$ : is a smoothing term that avoids division by zero.<br>Unlike other optimizers where the learning rate is constant for all parameters and for each cycle, Adagrad has parameter specific learning rates, which are adapted relative to how frequently a weight gets modified during training the model. More the updates for a weight, smaller the learning rate. |
|---|---|---|
| 6 | AdaDelta Optimizer (Zeiler, 2012) | In AdaDelta optimizer each parameter w is changed using an algorithm given by:<br>Algorithm :<br>$E[g^2]_t = \gamma E[g^2]_{t-1} + (1 - \gamma) g_t^2$<br>$w_{t+1} = w_t - \dfrac{\alpha}{\sqrt{E[g^2]_t + \varepsilon}}$<br>Where t : Time step ; $g_t$ : Sum of the squares of the past Gradients with respect to all weights ; $\varepsilon$ : is a smoothing term that avoids division by zero.<br>AdaDelta is a robust extension of Adagrad. It adapts learning rates based on a moving window of gradient updates, instead of accumulating all past gradients, this way it continues to learn even when many updates have been done. |
| 7 | RMSprop optimizer (Hinton, Srivastava & Swersky, 2012) | In RMSprop or Root Mean Square Propagation optimizer each parameter w is changed using an algorithm given by:<br>Algorithm : $w_{t+1} = w_t - \dfrac{\alpha}{\sqrt{(1-\gamma) g_{t-1}^2 + \gamma g_t + \varepsilon}} \cdot g_t$<br>Where t : Time step ; $g_t$ : moving average of square gradients ; $\varepsilon$ : is a smoothing term that avoids division by zero; $\gamma$ : is the decay term.<br>RMSprop tries to resolve Adagrad's radically diminishing learning rates by using a moving average of the squared gradient. |

6. The weights are slightly increased or decreased in order to achieve the goal of loss score minimization. This continuous adjustment of weights is called a training loop, which is iterated sufficient number of times, in order to achieve the goal.

## MAJOR TYPES OF NEURAL NETWORKS

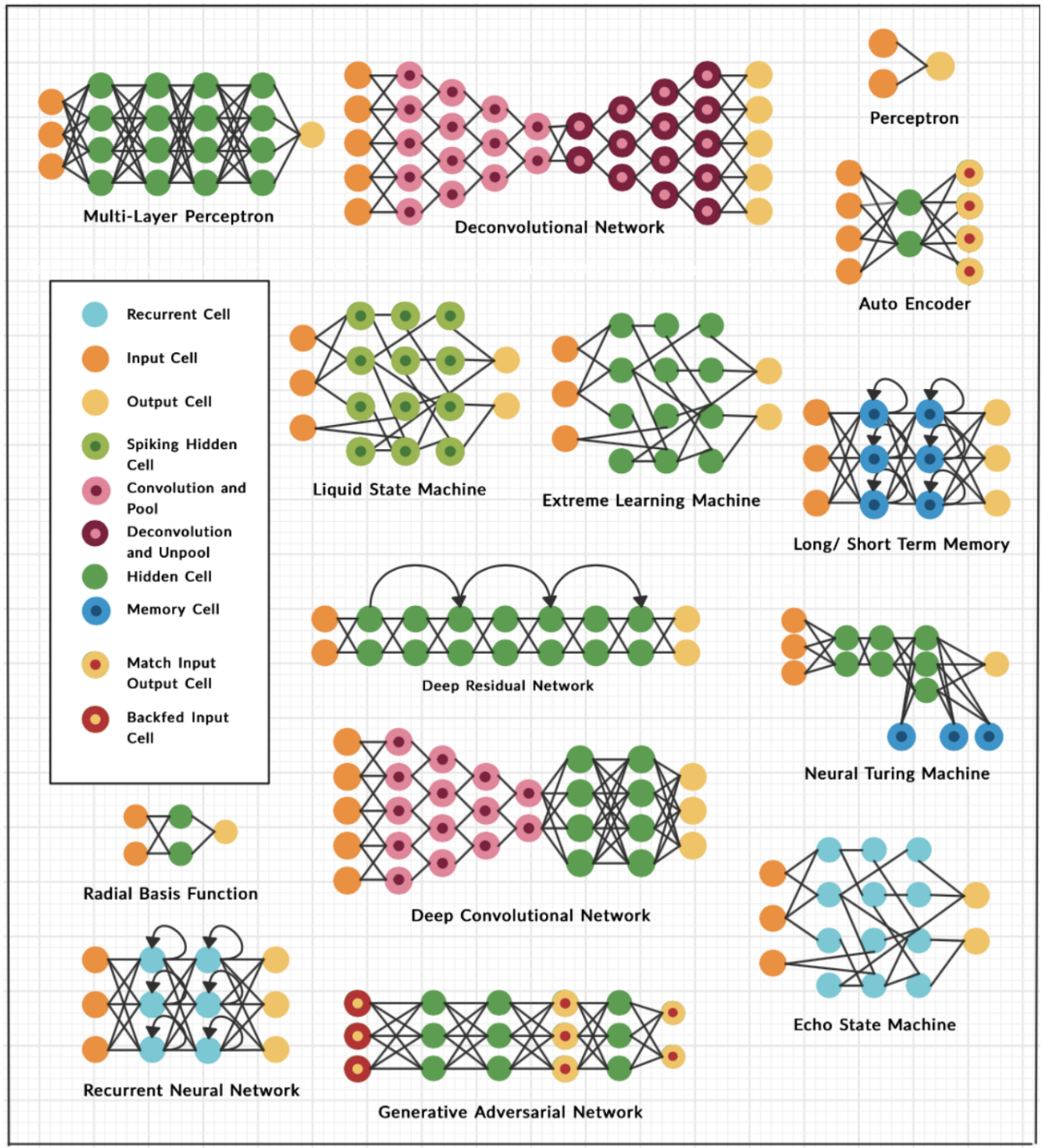

*Figure 3. Major Types of Neural Networks*

1. **Multi-Layer Perceptron**

A perceptron, by (Rosenblatt, 1958), is used for binary classification, and comes under the category of supervised machine learning. The elements of the training set are processed and learnt one at a time. Only linearly separable patterns are possible and are learned by a single layer perceptron. The linear decision boundary is drawn by learning the weights for the input signals, which makes the classes distinguishable. The simplest multi layer perceptron consists of an input layer, a hidden layer and an output layer. The input data is fed into the input layer and output is received from the output layer. It accepts the inputs, multiply them with suitable weights and moderate them, then pass them through the activation function to output the resultant value. The output gives a class score for the input and predicts the class of the input. The loss function is used to measure the performance of the predictor. High value of loss function indicates a larger difference between the predicted value and the actual value. The problems of overfitting and underfitting can be kept in check using an optimizer, thus a loss function and an optimizer are important for the optimization of the network. Usually, a common method is to assign random weights and to change these values iteratively to achieve lower loss function values. Figure 4 depicts the basic structure of a single layer perceptron and a multi layer perceptron.

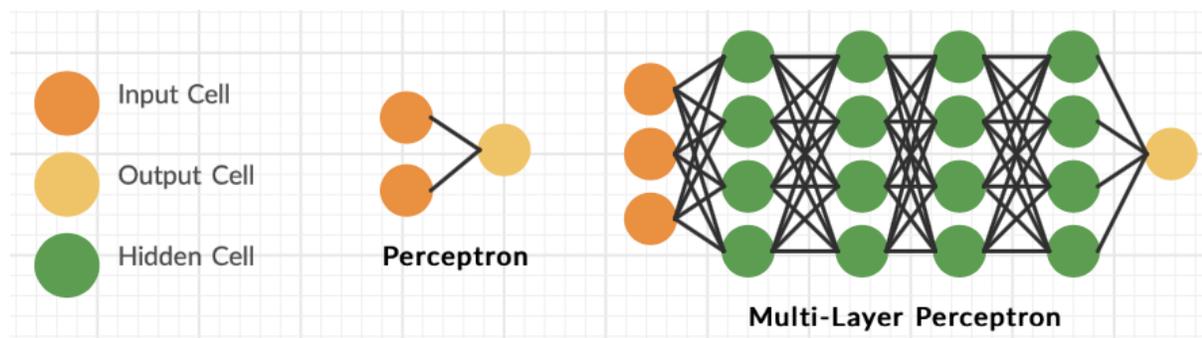

*Figure 4. Perceptron and Multi-Layer Perceptron*

The training of the network takes place in three steps :

- Forward pass : This step consists of feeding the network with the input values and, assigning and multiplying the inputs with the assigned weights. A bias value is added to each layer of the network and it is passed into the activation function to calculate the output.

- Calculating loss : The output obtained by passing the data instance is called the predicted value. The loss is calculated by calculating the difference between the predicted value of the instance and the actual value of the data instance.

- Backward pass : After the calculation of the loss, it is passed back into the network, using backpropagation, to update the values of the weights using gradient. The gradient flow determines the weights, which helps in determining the accuracy of the predictor.

    Major applications of MLP are Fitness approximation, universal function approximator, speech recognition, image recognition and in translation softwares.

2. **Radial Basis Function Neural Network**

Radial Basis Function (RBF) Neural Networks by (Broomhead & Lowe, 1988), employs the radial basis function as its activation function, which is a gaussian function, and are conceptually akin to the K-Nearest Neighbor algorithm, thus, the underlying concept is that, that the predicted target value of an item is likely to be the same as other items that have values close to the predictor values. The RBF network consists of three layers, which are, the input layer, the hidden layer and the output layer, as shown in Figure 5. The output of this type of neural network is always a real value, thus, it is used for function approximation. The yield of this network is a linear combination of neural parameters and inputs.

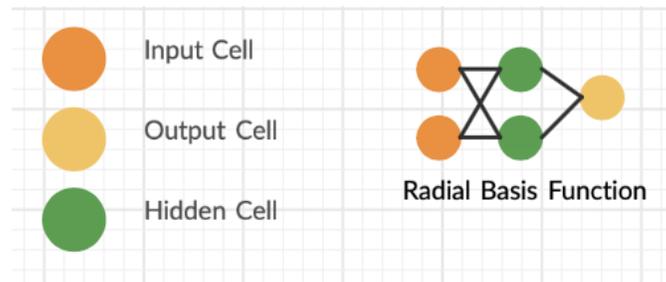

*Figure 5. Radial Basis Function Network*

Therefore, it can be said that a conventional RBF network has a linear output layer and a hidden layer containing a non-linear activation function. The optimization function's convergence is much faster as it contains only a single hidden layer, which qualifies RBF network to be the universal approximator. The RBF network finds vast applications like classification, interpolation and function approximation, in multiple fields and industries like stock price predictors, fraud detection and detection of anomalies in data and finances.

3. **Extreme Learning Machine**

Extreme Learning Machines or ELM by (Huang, 2015) are an important up-and-coming learning technique. The prime characteristics of this technique is that it does not need a learning process to compute or calculate the weights of a model. The main challenges faced when using popular machine learning techniques like SVM and NNs are: First, these techniques require intensive human intervention and Second, the learning speed is slow and Third, learning scalability is poor. So, good thing about ELM is that it achieves High Accuracy (for some cases), Least User Intervention and Real-Time Learning. The ELM not only achieves state of the art results but also lessens the time for training from days (spent by deep learning) to several minutes.

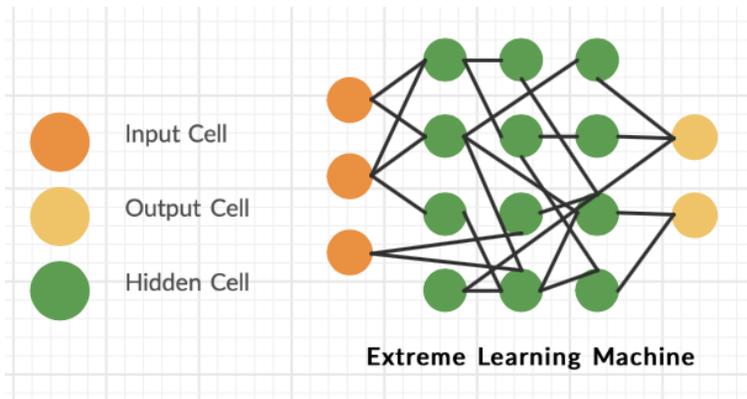

*Figure 6. Extreme Learning Machine*

The abridgment of ELM is as follows:
1. Architecture: ELM is a homogenous hierarchical learning machine for densely or partially connected layers of networks with almost any random number or type of hidden nodes.
2. Concept behind learning: ELM can be made to learn without iteratively tuning hidden nodes.
3. Learning theories : In ELM, learning is made layer wise. The hidden neurons are randomly generated or inherited from ancestors and the parameters of the model are learned faster.

The network architecture for ELM is given in Figure 6, where it can be seen that a random number of neurons are connected to any random numbers of neurons. As we move forward in the network, the top layers inherit any number of nodes from previous layers and each layer can have any random number of hidden nodes. The applications of ELM are Object detection, Image processing and speech recognition.

### 4. Recurrent Neural Network

The major characteristic of all neural networks like MLPs or Convnets, is that these networks have no memory. These networks keep no state between inputs fed to them and process each one of them independently. Whereas RNN or Recurrent Neural Network, created by (Hopfield, 1982), processes inputs fed to it incrementally while maintaining an internal model of what it is processing, this model is built from past experiences and is constantly updated with new information. RNN is one of a kind of NN with an internal loop. The diagram for simple RNN is given in Figure 7.

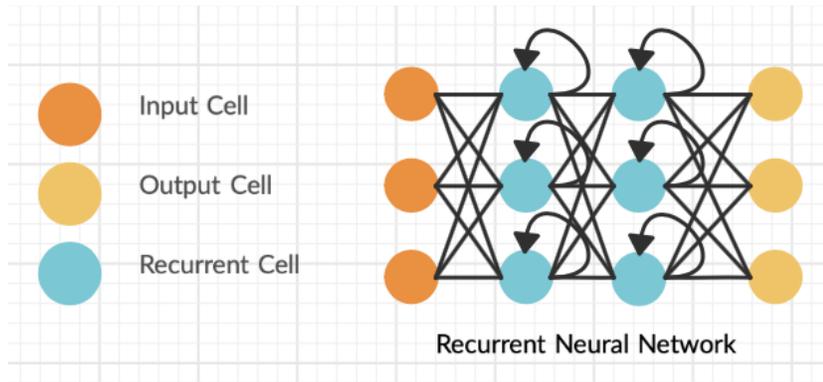

*Figure 7. Recurrent Neural Network*

The information is passed from one part of the network to the other using the loop. In addition to learning by training, it has the ability to remember prior information while generating new outputs. It acts as multiple copies of the same neural network cascaded together. These networks communicate with each other by passing messages. This unique structure gives RNN an edge over the normal perceptron and the convnets, by allowing operations over the sequences of inputs, outputs, or both. Let's consider a time $t$, so each layer in RNN combines the input fed to it at time $t$ with the output obtained by it at time $t-1$, in order to obtain output at time $t$. The two advanced forms of RNN are : LSTM, that is, Long Short-Term Memory and GRU, that is, Gated Recurrent Units.

Major applications of RNNs are document classification and time series classification (for eg. Identifying the topic of an article), time-series comparison (for eg. Estimating similarity between two documents), Language translation, Sentiment Analysis (for eg. Classifying the sentiment of tweets) and Weather forecasting.

### 5. Echo State Machine

Echo state network (ESN) or machine, created by (Jaeger, 2001), is a type of recurrent neural network with thinly distributed connections within the hidden layer, made up of recurrent cells. The weights and connectivity of the hidden and input layers are fixed, randomly assigned and untrainable. Whereas, the weights of the output cells are learned to produce temporal patterns. The idea behind the development of ESN was to enrich it with the capabilities of RNN but without the problems associated with it, like the vanishing gradient problem. ESN has a large reservoir, that is the hidden layer, which are connected sparsely. Figure 8 depicts a simple echo state network.

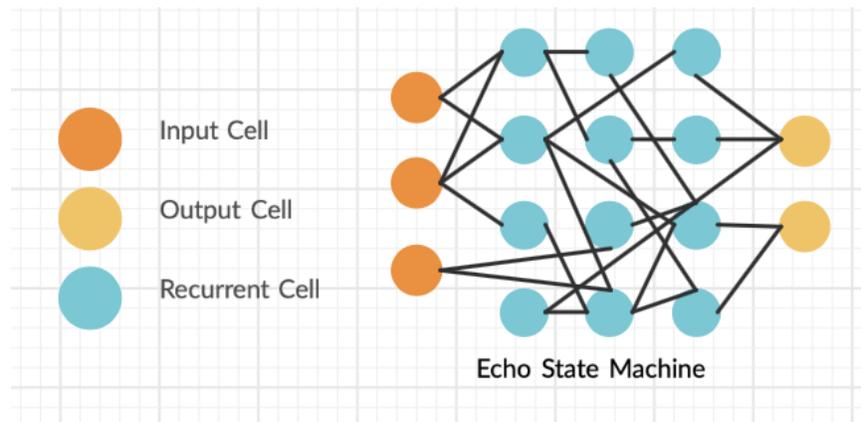

*Figure 8. Echo State Machine/ Network*

Sigmoid transfer function is used for the connection of the network's reservoir. The states of the reservoir are stored over time, and after all training inputs are completed, linear regression can be applied between the target outputs and the stored outputs of the reservoir. These resulting weights can be used in the existing network. The name comes from the idea that the previous states 'echo' with the help of the reservoir's random and sparse connections. When the network receives a novel input, tries to follow the activation trajectory of the similar input it trained on previously. This is a simple training procedure as the weights are assigned randomly and are untrainable. ESNs are used in signal processing applications, artificial soft limbs, optical microchips and polymer mixtures.

6. **Liquid State Machine**

Liquid State Machine or LSM, created by (Maass, 2010), is a particular kind of spiking neural network and is a complex topic that deals with topics like physics, chaos, dynamic action system, feedback system etc. LSMs was developed to explain the arcane operations of the brain. LSMs are considered to be an improvement over the theory of artificial neural networks because circuits in the network are not hard coded to perform a particular task and all continuous time inputs are handled naturally by the network and the same LSM network can perform multiple computations. In LSM, the control over the process of learning is minimal. The network architecture of LSM is given in Figure 9.

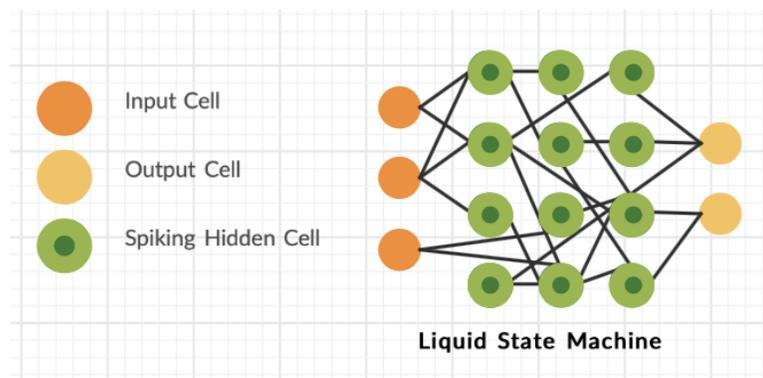

*Figure 9. Liquid State Machine*

The components are as follows :

1. Input Layer : Each node in this layer receives time varying inputs for the network from external sources.

2. Hidden layers: These layers consist of recurrent nodes and the nodes are randomly connected to each other. The time varying input forwarded by input layer is converted into a spatiotemporal pattern of activations by recurrent nature of connections in LSM. The concept of recurrently connected nodes in LSMs will end up computing a large variety of nonlinear functions on the input forwarded.
3. Output layer: The spatiotemporal patterns of activations forwarded by the network are read out by linear discriminant units present in the output layer.

The word liquid in the Liquid State Machine is an analogy for the chaotic ripples generated when we drop a stone into a still body of liquid. So, the time varying input from external sources (motion of falling stone) has been converted into patio temporal patterns of liquid displacement (the chaotic ripples). Applications of LSM are Speech recognition, object detection and Computer vision applications.

7. **Deconvolutional Neural Network**

The Deconvolution Neural Network or Deconvnet, created by (Noh, Hong & Han, 2015), is used to solve the task of semantic segmentation in the field of computer vision. The goal of semantic segmentation of image is to label each pixel of an image with a corresponding class of what is being represented and this task is very well performed by Deconvents. A Deconvnet takes an image processed by a convnet, as an input and outputs a high resolution image. In its architecture instead of pooling layers, it has Unpooling layers which increases the size of input as it moves through the Deconvnet circuit. A simple deconvolutional network is shown in Figure 10.

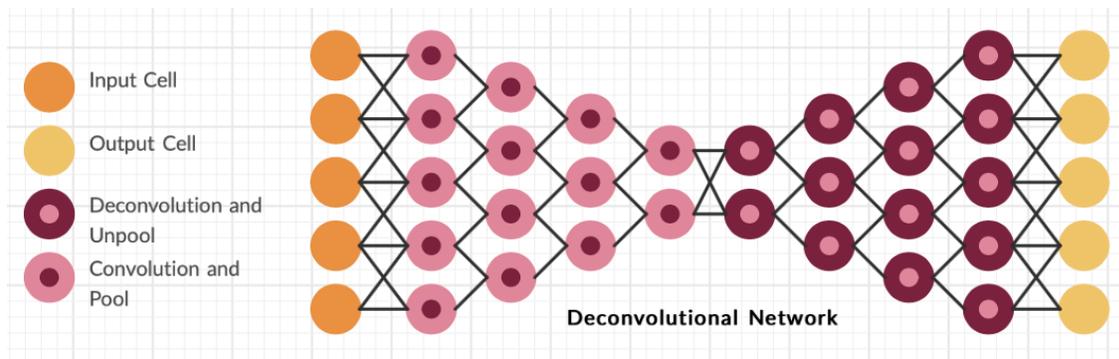

*Figure 10. Deconvolutional Network*

Some applications of deconvolution neural network are:
1. Autonomous vehicles : Deconvnets provide information about free space on the lands and detect lane traffic signs and markings.
2. Bio-medical image diagnosis : By using Deconvnets the time required to run a diagnostic test can be reduced significantly.
3. Geo sensing : Deconvnets are used by satellite imaging.
4. Precision agriculture : Deconvents provide precision to the drones used for agriculture usage like for Drone for irrigation.

8. **Deep Residual Network**

A deep residual network, commonly known as a deep ResNet, created by (He, Zhang, Ren & Sun, 2015), is a network with many layers. It builds the network using constructs known as pyramidal cells, and it does so by skipping connections, that is, jumping over several layers. This jumping over layers also leads to riddance of the 'vanishing gradient' problem otherwise faced by the network, due to the large number of layers. The problem of vanishing gradient is solved by using the activations from the previous layer again, till the layer adjacent to it learns the weight. Another problem caused by the large number of layers in the network is the degradation or saturation in the accuracy value of the results. The skip connections are responsible for adding the output of the previous layers to the output of the stacked layers, which makes the training of the deeper networks possible. A simple deep residual network is depicted in Figure 11.

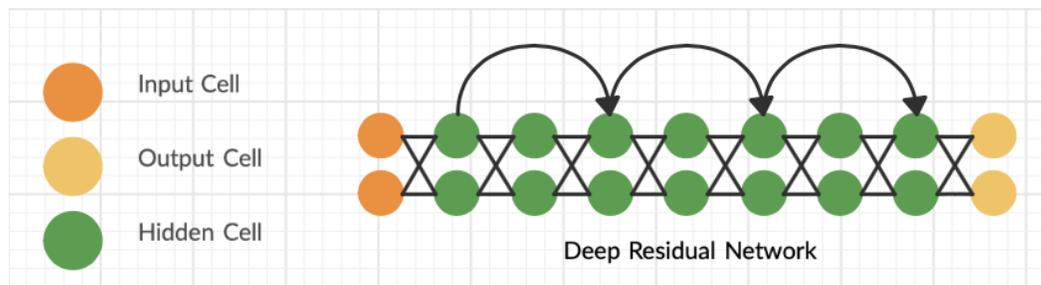

*Figure 11. Deep Residual Network*

It is also responsible for making the learning process faster than compared to the conventional neural networks, as propagation is through fewer layers. Deep residual networks are majorly used for image recognition, image reconstructions, Object detection, natural language processing and speech recognition.

9. **Deep Convolutional Network/Convolutional Network**

Convolutional Networks or Convnets, created by (LeCun, Bottou, Bengio & Haffner, 1998), are universally used in computer vision applications. The cardinal difference between Convnets and MLP is that, in MLP model learns global patterns in its input feature space and whereas Convnets learns local patterns. This key characteristic of Convnets gives it two thought-provoking properties:

- The patterns learnt by Convets are translation invariant : A Convnet, after learning a certain pattern in the upper-right corner of a picture, it can recognize it anywhere : for example, in the upper or lower left corner. A MLP would have to learn the same pattern again if it appeared at a new location.

- Convets can learn spatial hierarchies of patterns : A first convolution layer in Convnet will learn minute local patterns such as edges and a second convolution layer will learn patterns which are made of features learned by the first layer and so on. The simple structure of a convolutional network is shown in the figure below, that is figure 12.

The basic architecture of Convnet is as follows:
- Input Layer : This layer has N number of input cells which takes input for the network.

- Convolution Layer : This layer consists of convolution nodes, and each convolution node has its own activation function and a feature extraction matrix called the Kernel. Each node slides its kernel over the inputted image and performs matrix multiplication between the kernel and each portion of the image over which the kernel hovers.

- Pool layer : This layer is used for reduction of the spatial size (means dimensionality reduction) of the extracted convolved feature, thus decreasing the computation required to process the data. There are two types of pooling: Max pooling and Average pooling

- MLP : After extracting all the desired features, the extracted features are fed into a multi-layer perceptron for classification or regression.

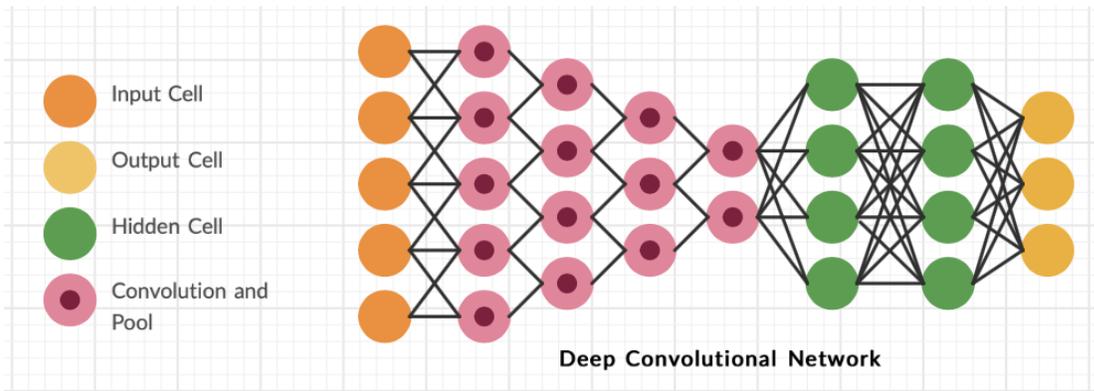

*Figure 12. Deep Convolutional Network*

In short, convnets used in the field of computer vision starts with a series of pooling and convolution layers, and ends with an MLP classifier. There are various pre-trained convnets, which are trained on very large data sets like ImageNet (Deng, Dong, Socher, Li, Li & Fei-Fei, 2009). These pre-trained convents can be used for solving any real-world problems. The various major types are given in table below :

*Table 4. The names of various convnets and their inventors*

| S.No. | Name of pertained Convnet | Name of inventor(s) |
| --- | --- | --- |
| 1 | VGG16 and VGG19 (Simonyan & Zisserman, 2014) | Karen Simonyan and Andrew Zisserman |
| 2 | ResNet (He, Zhang, Ren & Sun, 2015) | Kaiming He, Xiangyu Zhang, Shaoqing Ren and Jian Sun |
| 3 | Inception V3 (Szegedy, Liu, Jia, Sermanet, Reed, Anguelov, Erhan, Vanhoucke & Rabinovich, 2014) | Christian Szegedy, Wei Liu, Yangqing Jia, Pierre Sermanet, Scott Reed, Dragomir Anguelov, Dumitru Erhan, Vincent Vanhoucke and Andrew Rabinovich |
| 4 | MobileNets (Howard, Zhu, Chen, Kalenichenko, Wang, Weyand, Andreetto & Adam, 2017) | Andrew G. Howard, Menglong Zhu, Bo Chen, Dmitry Kalenichenko, Weijun Wang, Tobias Weyand, Marco Andreetto and Hartwig Adam |
| 5 | Mask R-CNN (He, Gkioxari, Dollár & Girshick, 2017) | Kaiming He, Georgia Gkioxari, Piotr Dollár and Ross Girshick |
| 6 | YOLO : You Look Only Once (Redmon, Divvala, Girshick & Farhadi, 2015) | Joseph Redmon, Santosh Divvala, Ross Girshick and Ali Farhadi |
| 7 | YOLO9000 (Redmon & Farhadi, 2016) | Joseph Redmon and Ali Farhadi |
| 8 | YOLOv3 (Redmon & Farhadi, 2018) | Joseph Redmon and Ali Farhadi |

## 10. Neural Turing Machine

Neural Turing machine (NTM) works on the concept of the availability of external memory in addition to the functionalities of the neural networks. It was formulated by Alex Graves (Graves, 2014). In layman terms, NTM comprises of two components, namely a memory device and a controller. The memory device or bank stores the inputs and outputs of the cell, which is controlled by the controller. The data can either be 'written to' or 'read from' the memory using parameter heads, which are used to access the memory. The controller acts as a link between the network's layers and the memory layer, and it allows and restricts the parameter heads to access the memory. The major steps involved in the NTM cells are :

- Taking input vector, let's say, X.
- Using this input vector X to find the output vector Y.
- Creating the read and write heads using the output vector Y.
- Computing the weights using these read and write heads.
- Read and write heads performing read and write operations respectively.

The read and write heads are used to compute the weights, which are responsible for the accessing of memory. A simple structure of the neural turing machine is depicted in Figure 13.

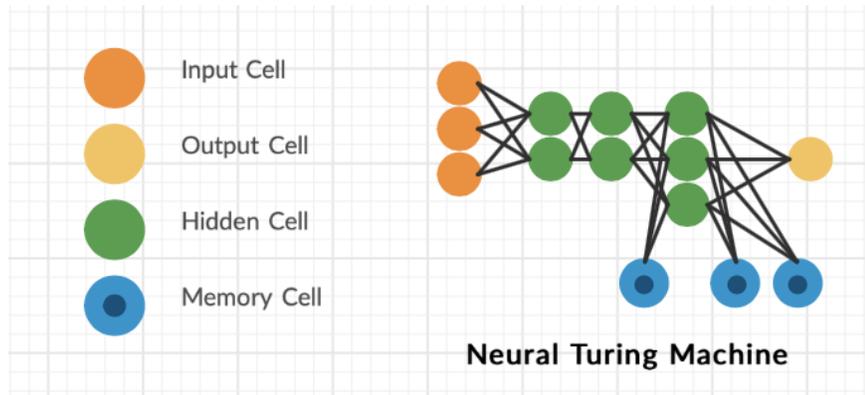

*Figure 13. Neural Turing Machine*

These weights are used to access memory as read and write heads cannot be discrete, that is, the read and write heads cannot tend to any single memory location. Thus, these heads are made to attend to different locations of the memory at different degrees, which is decided using the weights. To compute the weight vector, the head takes the controller output vector Y, the memory and the previous read/write vectors, as inputs. The controller extracts features from the inputs, and uses it to find the weights. The similarity between the extracted feature and memory entry, also called the K-score, is computed using the cosine similarity. The Softmax function is applied to this k-score to calculate the weight. It comprises of input cells, hidden cells, the memory cells and the output cells. Neural turing machine works well for learning small algorithms for detection of simple formal languages, copying and repeating. They can also be used for language modelling and balancing parentheses.

11. **Long/Short Term Memory Network**

Long/Short Term Memory networks were created by (Hochreiter & Schmidhuber, 1997) to solve the short-term memory problem of Recurrent Neural Networks (RNN), that is RNNs tend to leave important information from the beginning while processing a long enough sequence. This happens due to the vanishing gradient problem during back propagation in RNNs. The vanishing gradient problem takes place as the gradient goes through time and shrinks, and as it is used to update the values of the weights of the network, when the gradient value becomes too small, it has no effect on the changing net weight and eventually the learning process stops. As this happens in the early layers with smaller gradient values, the network's early layers stop learning and thus the RNNs can start to forget, thus inducing short-term memory problems. A simple LSTM network is depicted in Figure 14.

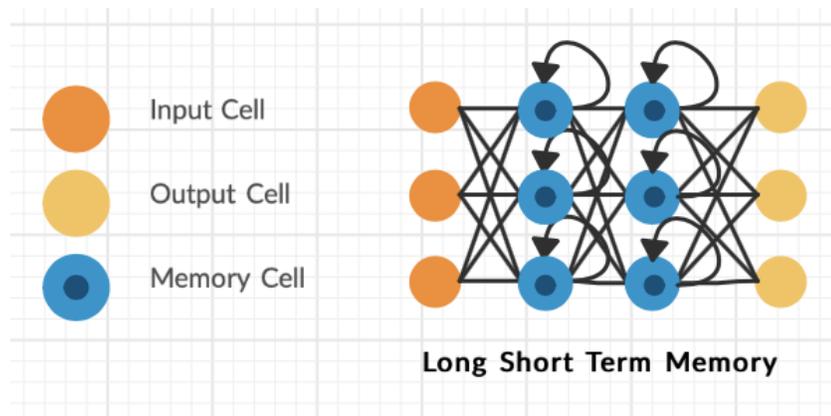

*Figure 14. Long Short Term Memory Network*

The LSTM network processes data or information as it passes through it in the forward direction. Here, the LSTM cell operations determine which information to keep, that is, to store and which information to forget. The main components of the LSTM network are:

- Cell State or memory : The cell state acts as the passage that allows or transfers desired information down the sequence chain. It acts as the memory component of the network. Information is added or removed from the memory as the cell passes through, with the help of the gates in the network. The gates learn to segregate the information into important and unimportant using the sigmoid function, and accordingly keeps or forgets the passed information during the training.

- The Forget gate : This gate chooses which information to keep and which to forget using the sigmoid function. As the sigmoid function gives output between 0 and 1, the information from current input and previous hidden state are passed in the signed function. If the value obtained is closer to 0 than 1, the information is forgotten, and if it is closer to 1, it is kept.

- The Input gate : The memory or cell state is updated using the input gate. It decides which information to update in the memory.

- The Output gate : The next hidden state is decided by the output gate. As the hidden states contain information of the previous inputs, they are vital for predictions.

Few real-world applications of LSTMs are short-term weather forecasting, robot control, speech recognition, music generation, handwriting recognition and semantic parsing.

12. Autoencoders

Autoencoders were created by (Hinton, Rumelhart & Williams, 1986). The output of these artificial neural networks is the reconstructed form of the input that is fed into it. These are unsupervised deep learning neural networks that work by compressing or encoding or reducing the dimensions of the input and then decode and reconstruct the compressed or encoded data into a very close representation of the original input passed. This process helps in removing the noise from the data by reducing the dimensions of the input data. Basic structure of an autoencoder is depicted in Figure 15.

An autoencoder mainly consists of four components :
- Encoder : This layer compresses the input data by reducing the dimensions of the input data. It changes the original input data into an encoded representation.
- Bottleneck : This layer has the maximum possible compressed form of the input data.

- Decoder : This layer is responsible for decoding the encoded representation of the input data. In this layer, the encoded data is decoded and the model learns to reconstruct the encoded data to be as close to the original input data.
- Reconstruction Loss : This is the measurement of how close the decoder output, that is, the reconstructed input, is to the original input. Backpropagation is used to minimize the value of reconstruction loss.

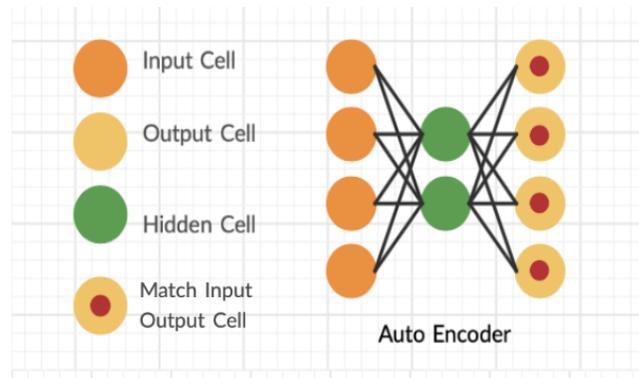

*Figure 15. Autoencoder*

The main practical applications of autoencoder include data denoising, that is the removal of noise in the data, and data dimensionality reduction. Denoising can be done for a signal, audio, image, document or for numeric data used for training models. There are mainly four different types of autoencoders used, they are :

- Vanilla autoencoder : This is the simplest form of autoencoder, with a total of three layers, namely, an input layer, a hidden layer and an output layer. The output is the reconstructed input, and it is done using a loss function.
- Multilayer autoencoder : This type of autoencoder consists of more than one hidden layer.
- Convolutional autoencoder : This type of autoencoder employs the use of convolutions instead of layers which are fully connected. They are mainly used for 3D images as input instead of the 1D vectors. It is a good method to downsample 3D images, by making the network learn from the dimensionally reduced and compressed form of the image.
- Regularized autoencoder : This type of autoencoder enables the model to introduce and enhance other properties in addition to the replication of the input to the output. This can be achieved by not keeping the encoder and decoder shallow. There are two types of regularized autoencoders, namely, sparse autoencoder and denoising autoencoder. Sparse autoencoder is used for the additional purpose of classification, as it learns features as a byproduct. Denoising autoencoder works by adding noise to the input image and making the autoencoder learn to remove it.

13. **Generative Adversarial Network**

Generative adversarial network (GAN), created by (Goodfellow, Pouget-Abadie, Mirza, Xu, Warde-Farley, Ozair, Courville & Bengio, 2014) highlights the concept of usage of deep learning for generative modeling, which is an unsupervised learning technique. It automatically discovers and learns the patterns and regularities in the data fed into it, because of which it is able to output or generate new but plausible patterns from the original data passed. GAN consists of two neural networks, called as the generator and the discriminator, competing with each other and thus gaining higher accuracy with each repetition or iteration. The job of the generator is to fabricate samples of data, which can be in the form of audio, image or numerical data, and the job of the discriminator is to distinguish between the actual and the fabricated data inputted. Thus, they are competing in a sense during the training phase. This helps both the networks to gain a deeper understanding of the data. The basic structure of a GAN is given in Figure 16. GANs are used for image editing, generating examples for image datasets, human faces, cartoons, realistic landscape pictures, image-to-text and text-to-image translation, 3D object generation and human pose generation.

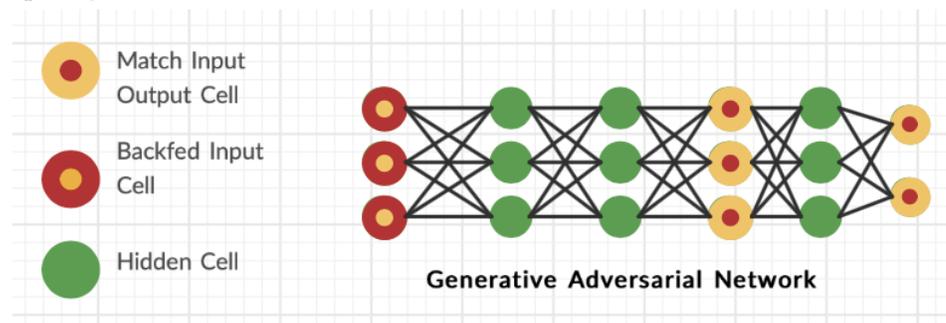

Figure 16. Generative Adversarial Network

The training of the GAN is made up of two phases, namely:
- Phase one : In this phase, training of the discriminator carried out without any activity in the generator. Backpropagation is not done and the network is trained only using forward propagation. The discriminator is trained on the real data and then on the fabricated data generated by the generator, to see if it can correctly predict them.

- Phase two : In this phase, training of generator is carried out with the discriminator being idle. The predictions made by the discriminator from the previous phase are used for the training of generator in this phase, so as to make it difficult for the discriminator to distinguish between the real and fabricated data samples.

The above process is carried out for a number of epochs. Manual checking of the fabricated data generated by the generator is also carried out, to make sure that the proper training is done. The training is stopped if the data generated seems real enough. There are five major types of GAN used currently, they are :
- Vanilla GAN : This type of GAN consists of, both the generator and the discriminator being simple multi-layer perceptrons and, is the simplest form of GAN available. Stochastic gradient descent is used for its optimization.

- Laplacian Pyramid GAN (LAPGAN) : This is a very high resolution image yielding approach, which employs the use of multiple generators and discriminators at different levels of the laplacian pyramid. In this, the image is first down-sampled and then up-scaled again to achieve higher resolution.

- Super Resolution GAN (SRGAN) : In this type of GAN, adversarial networks are used with deep neural networks to improve resolution of the image produced. SRGAN is majorly used for up-scaling the low-resolution images.

- Conditional GAN : In this type of GAN, an additional parameter is added to both the generator and the discriminator. A parameter 'y' is added to the generator for the fabrication of data, and a label is added to the input of the discriminator for the purpose of distinguishing the fake data samples from the real ones.

- Deep Convolutional GAN : This type of GAN is the most commonly used and the most successful type of GAN. In this GAN, in place of multi-layer perceptrons, the generators and discriminators consist of ConvNets without max pooling and the layers are also not fully connected.

# TABULAR SUMMARY

The details of the neural networks discussed in the previous sections have been neatly and briefly depicted in a tabular manner for the ease of understanding of the reader. The cogent description contains the name of the inventor and the year of invention of the respective neural network, along with their real-world applications.

*Table 5. The names of the inventors, year of invention and applications of the enlisted neural networks*

| Name of the network | Name of the inventor(s) | Year | Applications |
| --- | --- | --- | --- |
| Multi-layer perceptron | Frank Rosenblatt | 1958 | Fitness approximation, universal function approximator, speech recognition, image recognition, translation softwares |
| Recurrent neural network | J. J. Hopfield | 1982 | Signal forecasting, generation of text, music, Brain-Computer Interfaces, memory modeling, sentiment analysis |
| Autoencoders | David E. Rumelhart, Geoffrey E. Hinton & Ronald J. Williams | 1986 | Dimensionality Reduction, denoising of image/signal/audio /data |
| Radial basis function network | D. S. Broomhead & David Lowe | 1988 | System control & classification, time series prediction, function approximation |
| Long short term memory network | Sepp Hochreiter & JüRgen Schmidhuber | 1997 | Robot control, text generator as it can learn grammar, music composition, sign language translation, speech recognition |
| Deep convolutional network | Yann LeCun, Leon Bottou, Yoshua Bengio & Patrick Haner | 1998 | Object detection, video analysis, natural language processing, drug discovery, anomaly detection |
| Echo state machine | Herbert Jaeger | 2001 | signal processing, artificial soft limbs, optical microchips, polymer mixtures |
| Liquid state machine | Wolfgang Maass | 2010 | Speech recognition, object detection, computer vision applications |
| Generative adversarial network | Ian Goodfellow | 2014 | image editing, generating examples for image datasets, human faces, cartoons, realistic landscape pictures, image-to-text and text-to-image translation |
| Neural turing machine | Alex Graves | 2014 | learning and copying languages, language modelling |
| Extreme learning machine | Gao Huang | 2015 | Object detection, Image processing, speech recognition |
| Deconvolutional network | Hyeonwoo Noh, Seunghoon Hong & Bohyung Han | 2015 | Automated vehicles, medical diagnosis, geo-sensing, drones for agriculture |
| Deep residual network | Kaiming He, Xiangyu Zhang, Shaoqing Ren & Jian Sun | 2015 | Object detection, natural language processing, image generation, speech recognition |

## CONCLUSION

Neural networks are an abyss containing boundless opportunities for newer and more advanced inventions. The struggle to achieve thorough human-like intelligent behavior is an ongoing endeavor and is likely to remain one in the future as well. Even though the scientists and researchers have come a long way from where they started and have achieved astounding breakthroughs with image recognition, speech recognition, object detection, and predictive medical diagnostic tools, the goal is yet far-flung. And though neural networks hold more immanent potential than any other machine learning algorithm to build more customized and intelligent machines, it is very easy to get lost in their complexity. By means of this chapter, the authors are able to underscore the theory and concepts backing up the neural networks, their evolution- the basic idea behind their discovery, the structure and complexity of the conventional neural network. The chapter elegantly delineates the core theoretical ideas that make up the neural networks, the activation functions, loss functions and the optimizers used within. The mathematical equations and graphs of the same are also presented in a candid yet coherent manner, to augment the readers' grasp on the topic. The major types of neural networks discovered so far, the latest research, along with their basic structure, various embodying parts, ideas and logic, their training processes and their major applications are also listed in a well-grounded manner in the subsequent sections. Some significant flaws in the concept of neural networks are their extreme intricacy, time-consuming creation, extreme dependency on the condition of the data fed into them as inputs, and the lack of transparency in the learning, as well as in the predicting phase. But their virtues outweigh their vices. Neural networks have a galore of potential waiting to be discovered and utilized in the future. This chapter is a medium to invoke innovation in the readers and is like a handbook for neural networks and the arduous concepts supporting them, to guide the readers who are new to this field of study and give them a comprehensive understanding of the topic and its future prospects.

## REFERENCES

Ackley, D.H., Hinton, G.E. and Sejnowski, T.J. (1985), A Learning Algorithm for Boltzmann Machines*. Cognitive Science, 9: 147-169. doi:10.1207/s15516709cog0901_7

Binary crossentropy loss function: Peltarion Platform. (n.d.). Retrieved from https://peltarion.com/knowledge-center/documentation/modeling-view/build-an-ai-model/loss-functions/binary-crossentropy

*Broomhead, D.S. & Lowe, D. (1988).* Radial basis functions, multi-variable functional interpolation and adaptive networks

Calimera, A., Macii, E., & Poncino, M. (2013). The Human Brain Project and neuromorphic computing. *Functional neurology*, *28*(3), 191–196. doi:10.11138/FNeur/2013.28.3.191

Categorical crossentropy loss function: Peltarion Platform. (n.d.). Retrieved from https://peltarion.com/knowledge-center/documentation/modeling-view/build-an-ai-model/loss-functions/categorical-crossentropy

Cauchy, A. (1847). Méthode générale pour la résolution des systemes d'équations simultanées. *Comp. Rend. Sci. Paris*, *25*(1847), 536-538.

Copeland, B. J. (2019, November 19). Artificial intelligence. Retrieved from https://www.britannica.com/technology/artificial-intelligence.

Deng, J., Dong, W., Socher, R., Li, L. J., Li, K., & Fei-Fei, L. (2009, June). Imagenet: A large-scale hierarchical image database. In *2009 IEEE conference on computer vision and pattern recognition* (pp. 248-255). Ieee.

Duchi, J., Hazan, E., & Singer, Y. (2011). Adaptive subgradient methods for online learning and stochastic optimization. *Journal of machine learning research, 12*(Jul), 2121-2159.

Goodfellow, I.J., Pouget-Abadie, J., Mirza, M., Xu, B., Warde-Farley, D., Ozair, S., Courville, A.C., & Bengio, Y. (2014). Generative Adversarial Nets. *NIPS*.

Graves, A., Wayne, G., & Danihelka, I. (2014). Neural Turing Machines. *ArXiv, abs/1410.5401*.

He, K., Gkioxari, G., Dollár, P., & Girshick, R.B. (2017). Mask R-CNN. *2017 IEEE International Conference on Computer Vision (ICCV)*, 2980-2988.

He, K., Zhang, X., Ren, S., & Sun, J. (2015). Deep Residual Learning for Image Recognition. *2016 IEEE Conference on Computer Vision and Pattern Recognition (CVPR)*, 770-778.

Hendrycks, D., & Gimpel, K. (2016). Gaussian Error Linear Units (GELUs).

Hinton, G., Srivastava, N., & Swersky, K. (2012). Neural networks for machine learning lecture 6a overview of mini-batch gradient descent. *Cited on, 14*(8).

Hochreiter, S., & Schmidhuber, J. (1997). Long Short-Term Memory. *Neural Computation, 9*, 1735-1780.

Hopfield, J.J. (1982). Neural networks and physical systems with emergent collective computational abilities. *Proceedings of the National Academy of Sciences of the United States of America, 79 8*, 2554-8 .

Howard, A.G., Zhu, M., Chen, B., Kalenichenko, D., Wang, W., Weyand, T., Andreetto, M., & Adam, H. (2017). MobileNets: Efficient Convolutional Neural Networks for Mobile Vision Applications. *ArXiv, abs/1704.04861*.

Huang, G. (2015). What are Extreme Learning Machines? Filling the Gap Between Frank Rosenblatt's Dream and John von Neumann's Puzzle. *Cognitive Computation, 7*, 263-278.

Hyperbolic function. (n.d.). Retrieved from https://en.wikipedia.org/wiki/Hyperbolic_function

Jaeger, H. (2001). The"echo state"approach to analysing and training recurrent neural networks.

*Kurenkov, A. (2015, December 24). A 'Brief' History of Neural Nets and Deep Learning. Retrieved from https://www.andreykurenkov.com/writing/ai/a-brief-history-of-neural-nets-and-deep-learning/.*

LeCun, Y., Bottou, L., Bengio, Y., & Haffner, P. (1998). Gradient-based learning applied to document recognition.

Maass, W.S. (2010). Liquid State Machines: Motivation, Theory, and Applications.

Max-pooling / Pooling. (n.d.). Retrieved from https://computersciencewiki.org/index.php/Max-pooling_/_Pooling

Mcculloch, W. S., & Pitts, W. (1943). A logical calculus of the ideas immanent in nervous activity. *The Bulletin of Mathematical Biophysics*, *5*(4), 115–133. doi: 10.1007/bf02478259

Mean absolute error. (n.d.). Retrieved from https://en.wikipedia.org/wiki/Mean_absolute_error

Mean absolute percentage error. (2019). Retrieved from https://en.wikipedia.org/wiki/Mean_absolute_percentage_error

Mean squared error. (n.d.). Retrieved from https://en.wikipedia.org/wiki/Mean_squared_error

Mean squared logarithmic error loss function: Peltarion. (n.d.). Retrieved from https://peltarion.com/knowledge-center/documentation/modeling-view/build-an-ai-model/loss-functions/mean-squared-logarithmic-error

Nair, V., & Hinton, G.E. (2010). Rectified Linear Units Improve Restricted Boltzmann Machines. *ICML*.

Nesterov, Y. (1983). A method for unconstrained convex minimization problem with the rate of convergence o(1/k^2).

Neural computation. (2019, June 16). Retrieved from https://en.wikipedia.org/wiki/Neural_computation.

Noh, H., Hong, S., & Han, B. (2015). Learning Deconvolution Network for Semantic Segmentation. *2015 IEEE International Conference on Computer Vision (ICCV)*, 1520-1528.

*Norman, J. (n.d.). Alan Turing's Contributions to Artificial Intelligence 7/1948 - 1950. Retrieved from* http://www.historyofinformation.com/detail.php?id=4289.

Qian, N. (1999). On the momentum term in gradient descent learning algorithms. *Neural networks*, *12*(1), 145-151.

Redmon, J., & Farhadi, A. (2016). YOLO9000: Better, Faster, Stronger. *2017 IEEE Conference on Computer Vision and Pattern Recognition (CVPR)*, 6517-6525.

Redmon, J., & Farhadi, A. (2018). YOLOv3: An Incremental Improvement. *ArXiv, abs/1804.02767*.

Redmon, J., Divvala, S.K., Girshick, R.B., & Farhadi, A. (2015). You Only Look Once: Unified, Real-Time Object Detection. *2016 IEEE Conference on Computer Vision and Pattern Recognition (CVPR)*, 779-788.

*Richard, N. (2018, September 4). The differences between Artificial and Biological Neural Networks. Retrieved from* https://towardsdatascience.com/the-differences-between-artificial-and-biological-neural-networks-a8b46db828b7.

Robbins, H., & Monro, S. (1951). A stochastic approximation method. *The annals of mathematical statistics*, 400-407.

Rosenblatt, F. (1958). The perceptron: A probabilistic model for information storage and organization in the brain. *Psychological Review*, *65*(6), 386–408. doi: 10.1037/h0042519

Rumelhart, D.E., Hinton, G.E., & Williams, R.J. (1986). Learning internal representations by error propagation.

Sabour, S., Frosst, N., & Hinton, G.E. (2017). Dynamic Routing Between Capsules. *ArXiv, abs/1710.09829*.

Schuster, M., & Paliwal, K.K. (1997). Bidirectional recurrent neural networks. *IEEE Trans. Signal Processing, 45*, 2673-2681.

Sigmoid function. (n.d.). Retrieved from https://en.wikipedia.org/wiki/Sigmoid_function

Simonyan, K., & Zisserman, A. (2014). Very Deep Convolutional Networks for Large-Scale Image Recognition. *CoRR, abs/1409.1556*.

Softmax function. (2019, November 29). Retrieved from https://en.wikipedia.org/wiki/Softmax_function

Step function. (n.d.). Retrieved from https://en.wikipedia.org/wiki/Step_function

Szegedy, C., Liu, W., Jia, Y., Sermanet, P., Reed, S., Anguelov, D., Erhan, D., Vanhoucke, V., & Rabinovich, A. (2014). Going deeper with convolutions. *2015 IEEE Conference on Computer Vision and Pattern Recognition (CVPR)*, 1-9.

Werbos, P. J. (1974). Beyond regression: new tools for prediction and analysis in the behavioral sciences.

Zeiler, M. D. (2012). Adadelta: an adaptive learning rate method. *arXiv preprint arXiv:1212.5701*.